\documentclass{article}

    \usepackage[final]{neurips_2021}

\usepackage[utf8]{inputenc} 
\usepackage[T1]{fontenc}    
\usepackage{hyperref}       
\usepackage{url}            
\usepackage{booktabs}       
\usepackage{amsfonts}       
\usepackage{nicefrac}       
\usepackage{microtype}      
\usepackage{xcolor}         
\usepackage{soul}
\usepackage{graphicx}
\usepackage{subcaption}
\usepackage{float}
\title{Learning to identify cracks on wind turbine blade surfaces using drone-based inspection images} 

\author{%
   Akshay Iyer \\
   SkySpecs Inc. \\
  \small{\texttt{akshay.iyer@skyspecs.com}}
  \And
 Linh Nguyen\\ 
  SkySpecs Inc.\\
  \small{\texttt{linh.nguyen@skyspecs.com}}
   \And
   Shweta Khushu\\
   SkySpecs Inc.  \\
 \small{\texttt{shweta.khushu@skyspecs.com}}
}

\begin{document}

\maketitle

\begin{abstract}

Wind energy is expected to be one of the leading ways to achieve the goals of the Paris Agreement but it in turn heavily depends on effective management of its operations and maintenance (O\&M) costs. Blade failures account for one-third of all O\&M costs thus making accurate detection of blade damages, especially cracks, very important for sustained operations and cost savings. Traditionally, damage inspection has been a completely manual process thus making it subjective, error-prone, and time-consuming. Hence in this work, we bring more objectivity, scalability, and repeatability in our damage inspection process, using deep learning, to miss fewer cracks. We build a deep learning model trained on a large dataset of blade damages, collected by our drone-based inspection, to correctly detect cracks. Our model is already in production and has processed more than a million damages with a recall of 0.96. We also focus on model interpretability using class activation maps to get a peek into the model workings. The model not only performs as good as human experts but also better in certain tricky cases. Thus, in this work, we aim to increase wind energy adoption by decreasing one of its major hurdles - the O\&M costs resulting from missing blade failures like cracks.
\end{abstract}

\section{Introduction}
\label{section_intro}

While the year 2020 saw one of the most stunning declines in global CO$_{2}$ emissions due to the pandemic \cite{iea2020}, the economic recovery post-pandemic is set to reverse 80\% of that drop and global energy demand is set to increase by 4.6\% in 2021 \cite{iea2021}. However, renewables remain the hope with projections of 30\% contributions to electricity generation - the highest ever since the industrial revolution \cite{iea2021}. Wind alone is expected to be the prominent energy source by 2050 but for that to happen requires significant technological advancements to reduce the costs of wind power \cite{irena}. O\&M costs account for 20-24\% of the total Levelized Cost of Energy (LCOE) of current wind systems \cite{tend, curves} with blade failures being the major contributor to the costs as the turbines operate in harsh environmental conditions and are made up of complex, expensive materials \cite{blade1, blade2}. Damages on wind turbine blades, especially cracks on the surface, could indicate severe internal structural damages \cite{fracture} and if left to grow can cause serious damages \cite{babu} to the blades. Additionally, missing cracks result in repair expenses ranging in multiple hundreds of thousands of dollars, making it highly important to catch these damages at the earliest to better operate the turbines and avoid heavy losses. \cite{review, amirat}

\begin{figure}[htbp]
  \centering
  \includegraphics[scale=0.15]{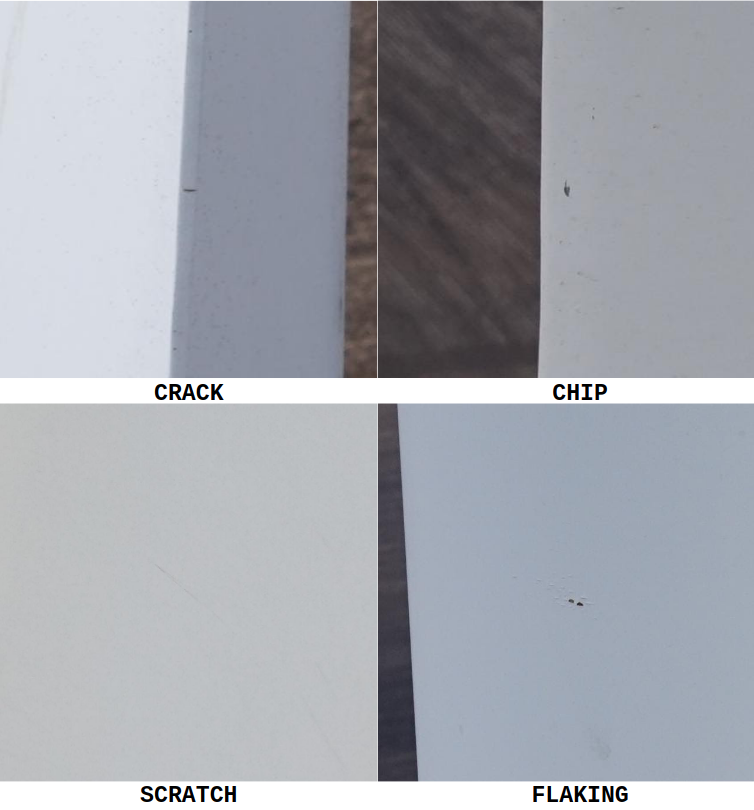}
  \caption{Blade damages - crack, chip, scratch, flaking. One can see how certain types of damages can be very faint and can sometimes look very similar to a crack}
  \label{fig_diff_dam_types}
  \vspace{-1.2em}
\end{figure}

There are a variety of sub-types in cracks like longitudinal, transverse, diagonal, buckling, etc., and can have very different visual appearances. At the same time, there are damage types other than cracks like chips, flaking, scratches, etc. which can at times look very similar to cracks but are not as severe as cracks. Thus, the problem becomes not only to correctly identify all the different sub-types of cracks but also to not confuse the other similar-looking damages to be cracks. Figure \ref{fig_diff_dam_types} shows how some other types of damages can look very similar to cracks and thus requires one to carefully discern the differences. Typically, this process is carried out manually making it very labor-heavy, subjective, and error-prone. This is where deep learning can improve upon human analysts and classical image processing methods. Since, with large enough data, deep learning models can identify discriminative, visual/non-visual features to distinguish cracks from the other very similar-looking damages and iteratively optimize that.  In this work, we focus on creating an accurate, fast, and reliable deep learning-based method to identify cracks on wind turbine blades. Our goal is to include deep learning as a part of our Quality Check (QC) process to miss fewer cracks.
In recent times, drone-based inspections of turbines have shown great promise as they avoid significant turbine downtimes and reduce the human risk in inspection\cite{drone1, drone2}. We perform drone-based inspections for wind farms capturing images in a variety of orientations, lighting, and weather conditions. From this, we carefully curate a dataset of 71k images and train a ResNet-50 on it, with specific training strategies. We then test the model on real-world data and take it to production. Our model is added as a layer in our QC process and the disagreements between the model and the human analyst are reviewed as a final step. 
\vspace{-1pt}

\par \textbf{Related Work}: 
There are some experimental works done in this domain like \cite{babu, haar, feasibility} use classical methods to detect certain cracks but can't distinguish them from similar-looking damages and are sensitive to noise and uneven illumination. Deep learning has garnered a lot of attention recently for tasks related to surface damages like on concrete \cite{concrete}, steel \cite{steel}, wood \cite{wood}, etc. But very limited work is done when it comes to this particular problem. Works like \cite{patel, reddy} use a CNN to classify if a blade image contains any kind of damage. However, they suffer from having just a few hundred images and resulting in false-positive-rates as high as 90\% \cite{patel}. \cite{vortex} considers a few categories but not cracks and achieves an mAP of 81\%. So most of these works either treat all damages the same or don't include cracks which have one of the most severe failure impacts. 
Thus, these works target related but slightly different aspects of the problem and also lack the data and surrounding infrastructure to get translated from an experimental stage to realize a real-world climate change impact. 


\label{section_contrib}
\par \textbf{Our contributions}: 
Our work is the first commercial product to create a crack-damage classifier trained on such a large amount of real-world data. Our model reports very high precision and recall on live production data and correctly identifies even very tricky cracks where trained analysts fail. We have laid special emphasis on the interpretability of results using Grad-CAM++ \cite{gradcam++} to increase trust in the predictions. The model runs very fast even on a CPU and unlike other works which are still in experimental stages, this one is already deployed to production bringing real-world value. Our model brings in repeatability, accuracy, and objectivity in our QC process. This would help human analysts make more informed decisions, miss fewer cracks, and result in better management of the turbines and their O\&M costs.



\section{Materials and Methods}
\label{section_methods}

\subsection{Data}
\label{section_data}
Our drones perform turbine inspections across the globe, capturing images of turbine blades in a variety of weather conditions thereby resulting in an information-rich dataset. So far we have performed around 90k inspections across 26 countries. The data then goes through a rigorous, manual QC process. In the initial round of QC, analysts identify damage locations, types, sub-types, and categorize them into five levels of severity based on factors like type, size, location, etc. of the damage. These annotations then go through a couple of QC rounds of review. We bring in our model at this stage as an \textit{auditor}, to give out its predictions on the images before they go to the last QC round where the experts would take the final call on the damages, especially cracks and high severity damages. Thus, our model becomes another round of QC to ensure that we miss fewer cracks. 

For this work, we curate a dataset of 71k images of cropped damages that have gone through our complete QC pipeline. We specifically focus on categories most often confused with cracks and vice versa by analysts, i.e. chips, flaking, scratch. The target labels were [\textit{'Crack', 'Not a crack'}]. We ensure to have a wide representation of the important sub-types of cracks. We perform a train-val split resulting in around 64k training images and 7k validation images. We perform data augmentation - both geometric and lighting-oriented to bring invariance to rotation and unfavorable lighting.

\subsection{Experiments}
\label{section_network_and_training}
\textbf{Network}: A ResNet50\cite{resnet} was initialized with ImageNet weights for the task and the last layer was replaced with a custom head consisting of fully connected layers, ReLU, and dropout for a binary prediction. An Adam optimizer was used with the learning rate found by using Leslie Smith's learning rate range test \cite{leslie} and a hyperparameter grid search. The network was implemented in PyTorch \cite{pytorch} and trained on a g4dn.xlarge GPU EC2 instance. 

\begin{figure*}[htbp]
  \begin{subfigure}[b]{0.5\linewidth}
    \centering
    \includegraphics[scale=0.26]{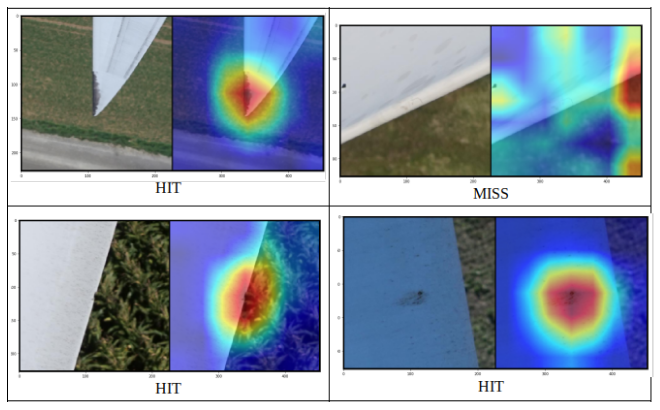}
    \caption{Grad-CAM++ visualization}
    \label{fig_gradcam}
  \end{subfigure}
  \begin{subfigure}[b]{0.5\linewidth}
    \centering
  \includegraphics[scale=0.15]{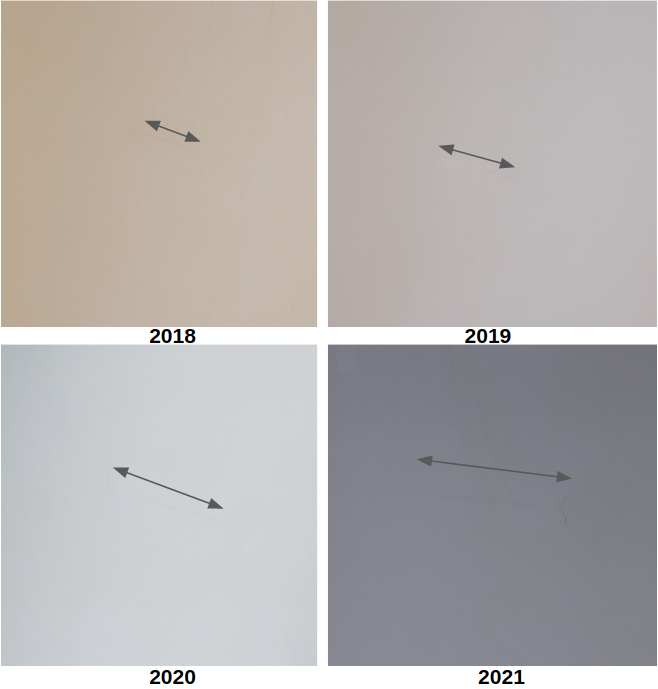}
    \caption{Propagation of cracks over time}
    \label{fig_propag}
  \end{subfigure}
  \caption{Results. Fig.\ref{fig_gradcam} shows Grad-CAM++ visualization for the model. The model mostly looks correctly at the damage (HITs) while making the predictions. Fig. \ref{fig_propag} shows how a very faint crack grew in size (shown with arrows) over four years. However, the model caught the crack in each case.}
  \label{fig_combined1}
\end{figure*}

\textbf{Training Strategies}: A weighted \textit{BCEWithLogits} loss was used to compensate for the class imbalance between cracks and no-cracks. An initial round of training from scratch was performed with the configuration. But since there are higher severity damages present in the data as well, which are more important to detect, we perform a second round of training to finetune the model to pay special attention to higher severity images. The higher severity damages (4 and 5) were present in a lesser proportion in the dataset (around 16\%). For this, we created a dataloader with custom stratified sampling to sample higher severity images much more such that they were not under-represented in batches. To take a peek into the workings of the network, Grad-CAM++ was used to generate a heatmap showing which parts of the input image were most responsible for the network prediction. The visualization and the interpretation can be seen in Fig. \ref{fig_gradcam}. While it is definitely most important to minimize Type II errors, it is also important not to have a large number of Type I errors else that shall result in a large number of unnecessary reviews in the QC process. Thus, it was most natural to use \textit{precision} and \textit{recall} as our machine learning metrics, and also since increasing these metrics imply fewer damages go undetected and propagate, thus directly corresponding with reduced O\&M costs.

\subsection{Evaluation}
\label{section_eval_and_deploy}
\textbf{Feedback loop with business}: To translate the ML performance to business value, there were several review rounds with the QC team. Here, the model was tested on images from production data i.e. real-world inspections. Then the class labels post the final round of QC were compared with the model predictions. This helped identify places where the model labeled incorrectly as well as cases when the model caught cracks that even the analyst missed. These catches along with the Grad-CAM++ visualization of the model helped the business trust the model. 

\begin{figure}[htbp]
  \centering
  \includegraphics[scale=0.17]{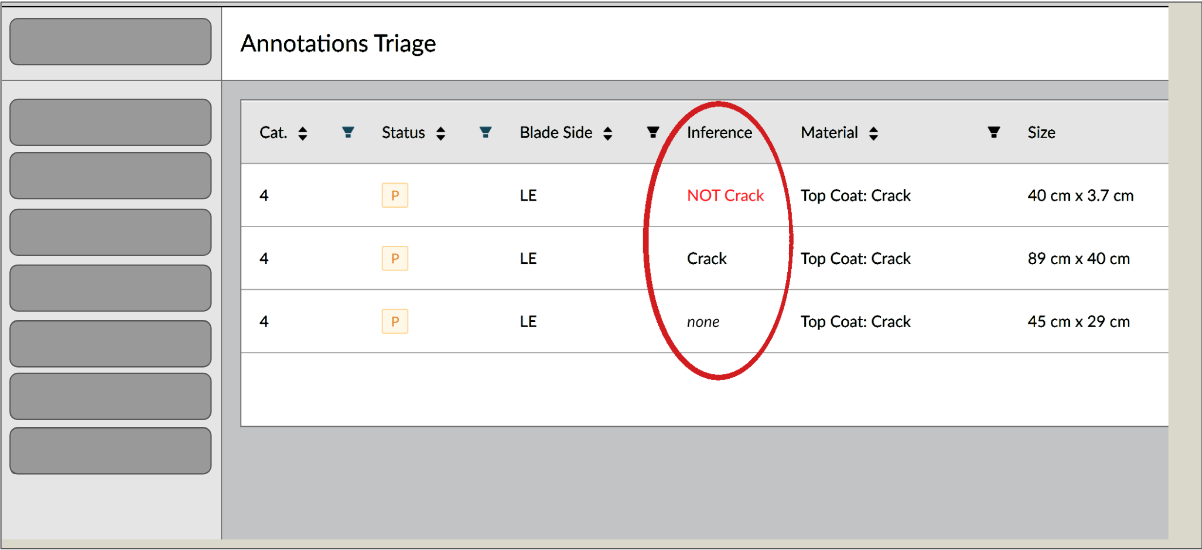}
  \caption{Model prediction on every damage visible to analysts to use in the QC process}
  \vspace{-1.2em}
  \label{fig_analyst_app}
\end{figure}

\textbf{Deployment Pipeline}:
Once the model reached the desired performance, model deployment began. For the same, the model was first optimized using the \textit{jit torchscript} compiler. The model is served using torchserve and hosted on a Sagemaker endpoint. The entire pipeline is built using AWS to ensure scalability and high availability. The model prediction is visible in the Analyst app used in the QC process for each image (see Fig. \ref{fig_analyst_app})



\begin{table}[htbp]
  \caption{Model performance on the test set and effect of including severity sampling. We observe the model has both high recall and precision. Also, severity sampling helps improve the performance}
  \label{table_results}
  \centering
  \scriptsize
  \begin{tabular}{lcccc}
    \toprule
    Data & Recall & Precision & F1-Score   \\
    \midrule
     Complete test data & $0.96$ & $0.85$ & $0.90$  \\
     Only high severity images (before severity sampling) & $0.92$ & $0.98$ & $0.95$  \\
     Only high severity images (after severity sampling) & $0.96$ & $0.98$ & $0.97$  \\
    \bottomrule
  \end{tabular}
\end{table}

\section{Results and Discussion}
\label{section_results}


Table \ref{table_results} shows how the model performs in production on around 46k images. It can be seen that on the complete test data, the model has a very high recall of 0.96 correctly identifying 23.2k cracks out of 24.2k cracks. The model results in not just very high recall but also high precision, thus not only helping analysts miss fewer cracks but also not adding to their burden of inspecting many false positives. Not only that, the model even caught several tricky cracks that even trained analysts had missed. So the model is not just performing similar to a human analyst, it has also started to perform better in some cases. Table \ref{table_results} also shows the effects of the custom sampling on the model performance on higher severity samples from a test set. Clearly, including stratified sampling for severity images helped the performance, the recall jumped four points while maintaining near-perfect precision. Tests revealed that this change was statistically significant with a p-value of 4e-7. Additionally, Fig. \ref{fig_propag}, shows a faint crack in 2018 which grew in size in successive years. On running the model, we found that the model caught it in all four years including the very first year while it still was a small and faint damage. Thus, the model catching such damages in very early stages will save hundreds of thousands of dollars spent on repairing larger cracks down the line. With more production data coming in and our pipeline in place to iteratively improve the model based on its failures, the performance can only increase. Also, to increase the trust in the model predictions, we used GradCAM++ for results interpretation. Fig. \ref{fig_gradcam} shows a heatmap overlaid on the image, with \textit{red} indicating regions where model focuses most and \textit{blue} indicating least. We observed that the model, without being explicitly taught, still focuses correctly on the damage to make the decision. This was used to increase trust in the predictions as well as debug model failures. Apart from accuracy, the model is also optimized for deployment, thereby yielding very fast inference times of 0.15 secs on a CPU (c4 instance). This enables us to have cheap and high availability. The model performs inspections on several thousand images within a few minutes which takes human analysts several hours. Fig.\ref{fig_analyst_app} shows how the model prediction is shown to the analysts for each image and gets updated in real-time. 

Thus, in this work, we create one of the first at-scale, in-production crack-damage classifiers to help reduce the O\&M costs of wind turbines and thereby increase wind energy adoption. The model, with its high scores and interpretability, brings in accuracy and reliability. It improves the damage inspection process thereby resulting in fewer missed cracks and potentially preventing significant expenses down the line.  However, there is still scope to increase accuracy which we plan to do by monitoring the model in production and using tools like Grad-CAM++ and curriculum learning \cite{curr} to investigate and improve the model. Also, the current focus is only on cracks but we are in talks with the business to identify important fine-grained sub-types of cracks. We eventually plan to combine this damage classifer with our other work on using deep learning to localize damages.\cite{blue}

\section{Acknowledgements}
Much of this work was made possible by the SkySpecs internal analyst team. We would also like to thank the anonymous reviewers for their insightful comments and feedback.

\setcitestyle{numbers}
\bibliography{refs}

\end{document}